\PassOptionsToPackage{table,svgnames}{xcolor}

\documentclass[sigconf]{acmart}

\usepackage{svg}
\usepackage{soul}
\usepackage{booktabs}
\usepackage{graphicx}
\usepackage{subfigure}
\usepackage{amsthm}
\usepackage{amsmath}
\usepackage{multirow}
\usepackage{enumerate}
\usepackage{amsfonts}
\usepackage{url}
\usepackage{tabularx}
\usepackage{diagbox}
\usepackage{longtable}
\usepackage{rotating}
\usepackage{slashbox}
\usepackage{algorithm}
\usepackage{algpseudocode}
\usepackage{bm}
\usepackage{float}
\usepackage{hyperref}
\usepackage[justification=centering]{caption}
\usepackage{pifont}

\definecolor{lightcoral}{RGB}{240,128,128}
\definecolor{lightblue}{RGB}{173,216,230}

\usepackage[normalem]{ulem}
\newcommand{\highlight}[2]{\sethlcolor{#1}\hl{#2}}
\newcommand{\cmark}{\ding{51}}%
\newcommand{\xmark}{\ding{55}}%
\def\BibTeX{{\rm B\kern-.05em{\sc i\kern-.025em b}\kern-.08em
    T\kern-.1667em\lower.7ex\hbox{E}\kern-.125emX}}
\AtBeginDocument{%
  \providecommand\BibTeX{{%
    Bib\TeX}}}

\setcopyright{acmlicensed}
\copyrightyear{2018}
\acmYear{2018}
\acmDOI{XXXXXXX.XXXXXXX}
\acmConference[Conference acronym 'XX]{Make sure to enter the correct
  conference title from your rights confirmation email}{June 03--05,
  2018}{Woodstock, NY}
\acmISBN{978-1-4503-XXXX-X/2018/06}

\copyrightyear{2025}
\acmYear{2025}
\setcopyright{acmlicensed}
\acmConference[KDD '25] {Proceedings of the 31st ACM SIGKDD Conference on Knowledge Discovery and Data Mining V.2}{August 3--7, 2025}{Toronto, ON, Canada.}
\acmBooktitle{Proceedings of the 31st ACM SIGKDD Conference on Knowledge Discovery and Data Mining V.2 (KDD '25), August 3--7, 2025, Toronto, ON, Canada}
\acmDOI{10.1145/3711896.3737198}
\acmISBN{979-8-4007-1454-2/25/08}

\begin{document}

\title{\textbf{Biological Pathway Guided Gene Selection Through Collaborative Reinforcement Learning}}

\author{Ehtesamul Azim}
\affiliation{%
  \institution{University of Central Florida}
  \city{Orlando}
  \state{Florida}
  \country{USA}
}\email{ehtesamul.azim@ucf.edu}

\author{Dongjie Wang}
\affiliation{%
  \institution{University of Kansas}
  \city{Lawrence}
  \state{Kansas}
  \country{USA}}
\email{wangdongjie@ku.edu}

\author{Tae Hyun Hwang}
\affiliation{%
  \institution{Vanderbilt University Medical Center}
  \city{Nashville}
  \state{Tennessee}
  \country{USA}}
\email{taehyun.hwang@vumc.org}

\author{Yanjie Fu}
\affiliation{%
 \institution{Arizona State University}
 \city{Tempe}
 \state{Arizona}
 \country{USA}}
\email{yanjie.fu@asu.edu}

\author{Wei Zhang$^\dagger$}
\affiliation{%
 \institution{University of Central Florida}
 \city{Orlando}
 \state{Florida}
 \country{USA}}
\email{wzhang.cs@ucf.edu}
\thanks{$\dagger$ Corresponding Author}

\renewcommand{\shortauthors}{Ehtesamul Azim et al.}

\begin{abstract}
    Gene selection in high-dimensional genomic data is essential for understanding disease mechanisms and improving therapeutic outcomes. Traditional feature selection methods effectively identify predictive genes but often ignore complex biological pathways and regulatory networks, leading to unstable and biologically irrelevant signatures. Prior approaches, such as Lasso-based methods and statistical filtering, either focus solely on individual gene-outcome associations or fail to capture pathway-level interactions, presenting a key challenge: how to integrate biological pathway knowledge while maintaining statistical rigor in gene selection? To address this gap, we propose a novel two-stage framework that integrates statistical selection with biological pathway knowledge using multi-agent reinforcement learning (MARL). First, we introduce a pathway-guided pre-filtering strategy that leverages multiple statistical methods alongside KEGG pathway information for initial dimensionality reduction. Next, for refined selection, we model genes as collaborative agents in a MARL framework, where each agent optimizes both predictive power and biological relevance. Our framework incorporates pathway knowledge through Graph Neural Network-based state representations, a reward mechanism combining prediction performance with gene centrality and pathway coverage, and collaborative learning strategies using shared memory and a centralized critic component. Extensive experiments on multiple gene expression datasets demonstrate that our approach significantly improves both prediction accuracy and biological interpretability compared to traditional methods.
    \let\thefootnote\relax\footnotetext{Release code and preprocessed data can be found at \url{https://github.com/ehtesam3154/bioMARL}}
\end{abstract}

\ccsdesc[500]{Applied computing~Computational genomics}

\ccsdesc[500]{Computing methodologies~Feature selection}
\ccsdesc[500]{Computing methodologies~Multi-agent reinforcement learning}

\keywords{automated gene selection; multi-agent reinforcement learning; biological pathway; disease outcome prediction}


\maketitle
\vspace{-0.25cm}
\section{Introduction}
With the rapid advancement of high-throughput technologies, high-dimension, low-sample-size (HDLSS) data have become increasingly prevalent in biomedical research. Large cohort cancer studies, such as The Cancer Genome Atlas (TCGA) \cite{cancer2013cancer}, contain vast amounts of HDLSS genomic data, offering an unprecedented opportunity to understand cancer mechanisms and improve therapeutic outcomes \cite{hartmaier2017high,ahmed2022multi,baul2022omicsgat}. However, extracting meaningful insights from this large genomic data requires addressing major analytical challenges, such as statistical randomness, experimental noise, and sample heterogeneity, which often result in inconsistent and biologically or clinically irrelevant gene signatures \cite{zhang2012signed}.

A well-known example of this challenge is the discrepancy between two early breast cancer diagnostic gene panels: MammaPrint, a 70-gene signature developed in the Netherlands \cite{van2002gene}, and a 76-gene signature from a similar study in San Diego \cite{wang2005gene}. Despite their shared goal of predicting disease progression, only three genes overlapped, highlighting the difficulty of selecting stable gene signatures (i.e., biomarkers). Traditional feature selection methods, such as Lasso-based approaches \cite{tibshirani1996regression,lemhadri2021lassonet}, have been widely used for biomarker selection due to their ability to enforce sparsity and identify predictive genes. However, these methods primarily focus on individual gene-outcome associations while neglecting the complex biological pathways and regulatory networks that govern gene function. Genes do not operate in isolation; rather, they interact within intricate systems, and integrating biological pathway information as prior knowledge can enhance the robustness and interpretability of selected gene signatures \cite{chuang2007network}. By leveraging known gene interactions and functional relationships, gene selection methods can improve reproducibility and biological relevance.

That said, integrating biological pathway knowledge into gene selection presents unique challenges that extend beyond simple statistical incorporation. While databases like KEGG \cite{kanehisa2000kegg} provide extensive biological pathway information, effectively leveraging this knowledge requires understanding complex dependencies between genes and their collective impact on biological processes. Most traditional feature selection methods operate independently of biological pathway knowledge, relying purely on statistical patterns in the data \cite{zhang2017network}. Simple filtering or scoring mechanisms that attempt to integrate pathway information often fail to capture the dynamic nature of gene interactions, leading to suboptimal gene selection. This highlights the need for a framework that models these complex dependencies while making biologically informed decisions about gene importance, ensuring selected gene signatures reflect mechanisms rather than purely statistical associations.

\textbf{Our contribution: A pathway-aware gene selection perspective.} \ul{We approach the gene selection problem from the perspective of collaborative decision-making, where each gene acts as an intelligent agent that learns when its inclusion benefits both predictive power and biological relevance.} The key insight is that: genes function through complex regulatory pathways and networks, interacting via molecular mechanisms, protein-protein interactions, and signaling cascades to drive cellular functions. Our selection agents mirror this biological reality by learning to make decisions that consider both their individual contributions and their collective impact within these pathways. By prioritizing genes with established pathway annotations, our framework leverages decades of experimental validation and enables more interpretable results, as selected genes can be understood within known biological processes. This biological context-aware selection also facilitates cross-study validation and increases the likelihood of identifying druggable targets, as existing therapeutics often target specific pathways.

This paper presents a two-stage architecture where pathway-guided statistical pre-filtering creates a biologically relevant reduced gene set, followed by a MARL-based selector where genes function as collaborative agents through sophisticated mechanisms including intelligent reward design, shared memory, and centralized feedback to maintain both statistical significance and biological relevance while optimizing predictive performance.
Our key contributions are as follows: 
\begin{itemize}
    \item[$\bullet$] First framework to unify statistical feature selection with pathway knowledge in genomic data analysis.
    \item[$\bullet$] Novel application of MARL to model gene selection as a collaborative decision process.
    \item[$\bullet$] Introduction of biologically-informed learning mechanisms that maintain both predictive power and pathway relevance.
\end{itemize}


\begin{figure*}[!h]
\vspace{-0.2cm}
    \centering
    \includegraphics[width=1.0\linewidth]{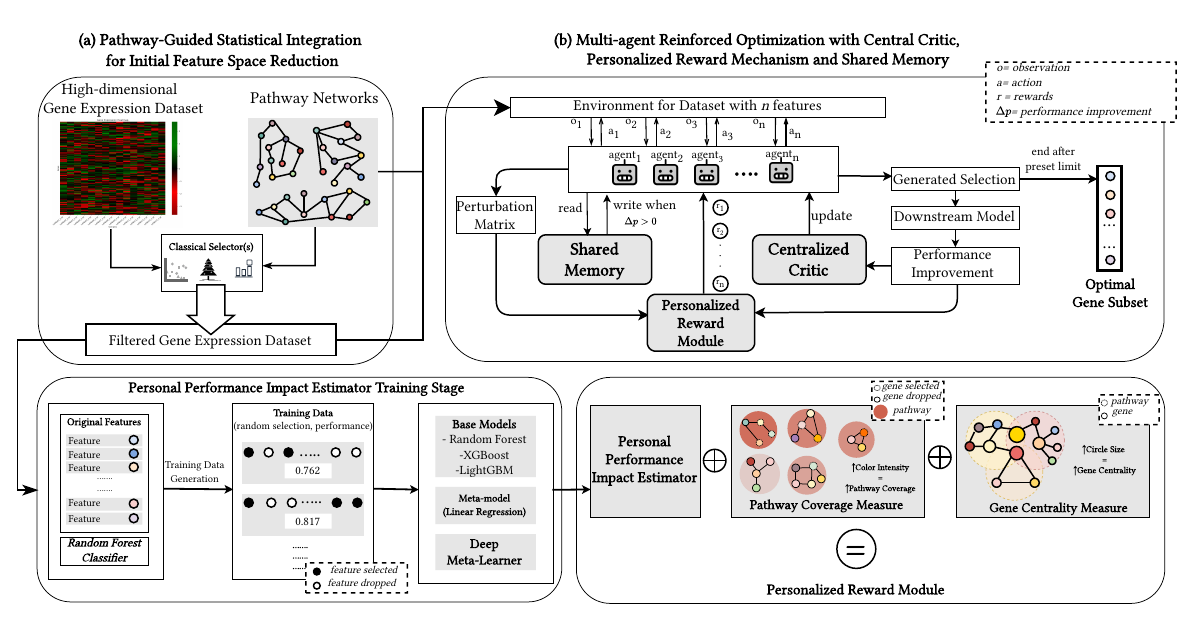}
    \vspace{-0.5cm}
    \captionsetup{justification=centering, font=small}
    \caption{
    Framework Overview. \textbf{BioMARL} consists of two key parts: (a) pathway-guided meta-selection combining multiple statistical methods with KEGG pathway information for initial filtering; (b) a pathway-aware multi-agent RL framework for refined selection, with centralized evaluation, shared memory and a multi-component reward system to enhance collaboration among agents while preserving biological relevance. The iteratively generated feature set is evaluated in downstream task, with iterations continuing until optimization or set limit.
    }
    \vspace{-0.5cm}
    \label{fig:framework}
\end{figure*}
\vspace{-0.2cm}
\section{Problem Statement}
This paper addresses the challenge of selecting relevant genes from high-dimensional genomic datasets. The goal is to identify a subset of genes that optimize predictive performance while maintaining biological interpretability and pathway-level coherence, addressing the limitations of traditional feature selection methods in capturing complex gene interactions. Formally, given a high-dimensional genomic dataset \( D \) with gene set \( G \), where each sample consists of gene expression values and an associated outcome, our objective is to select an optimal subset \( G_{opt} \subset G \) such that \( |G_{opt}| \ll |G| \), significantly reducing the dimensionality while preserving predictive power. Through a two-stage selection process, we first obtain a pre-filtered set \( G_{pre} \subset G \) using pathway-guided statistical methods, followed by a refined selection process that produces the final ranked set \( G_{ranked} \) from which the top \( k \) genes form \( G_{opt} \). The selected genes should not only optimize the performance of downstream prediction tasks but also reflect meaningful biological pathway relationships, ensuring both statistical significance and biological relevance in the final gene signature.
\vspace{-0.1cm}
\section{Methodology}
\subsection{Framework Overview}
\label{overview}
Figure \ref{fig:framework} illustrates our pathway-aware gene selection framework \textbf{BioMARL} consisting of two main components: The first component implements a biological pathway-guided gene pre-filtering that combines multiple statistical selection methods with pathway information from the KEGG database. This initial filtering creates a biologically relevant reduced gene set by evaluating both statistical significance and pathway-level performance. The filtered gene set then serves as input to our MARL based selector, where each gene is modeled as an independent agent within a collaborative framework. This incorporates pathway knowledge at multiple levels: through pathway-aware state representations using Graph Neural Networks and a reward mechanism that considers both statistical and pathway-based performance. The framework is further enhanced with collaborative learning mechanisms including a shared memory system for effective knowledge sharing between agents and a centralized critic for evaluating collective agent behavior.  This ultimately outputs a ranked list of selected genes that demonstrate both strong predictive power and biological relevance.

\vspace{-0.2cm}
\subsection{Pathway-Guided Gene Pre-filtering}
\label{stage_1}
\noindent\textbf{Why integrate biological context with statistical selection?} Purely statistical feature selection in genomic data analysis faces two fundamental challenges. First, the curse of dimensionality in HDLSS genomic datasets results in unstable feature\footnote{In this work, we use `feature' and `gene' interchangeably, as `feature' is the standard term in the broader feature selection literature, including gene selection studies.} rankings, where small variations in data can lead to substantially different selected gene sets. Second, complex non-linear interactions between genes make it difficult to accurately determine gene importance using individual statistical measures alone. Although advanced statistical techniques, such as regularization and ensemble methods, attempt to mitigate these issues, they remain biology-agnostic and fail to capture functional gene relationships. A biologically informed approach, leveraging pathway knowledge, introduces natural constraints and grouping structures to address these challenges. It reduces the effective search space by prioritizing biologically plausible feature combinations, while the inherent pathway-based organization of genes facilitates the identification of functionally relevant interactions that purely statistical approaches may overlook. By embedding pathway information, gene selection becomes more stable, interpretable, and computationally efficient, focusing on gene sets that align with known biological mechanisms rather than purely data-driven statistical associations. 
We implement a pathway-guided pre-filtering strategy that combines statistical feature selection with biological pathway knowledge. The pre-filtering process consists of three main components:

\noindent\textbf{\ul{\emph{Base Score Computation.}}}
Multiple statistical feature selection methods are employed to compute initial importance scores for each gene. For a dataset $D$ with gene set $G$, each method $f_i \in F$ generates a score vector $S_i$, where $S_{i,g}$ represents the importance score assigned to gene $g$ by method $f_i$. Our framework employs three methods ($|F| = 3$): chi-squared test, random forest importance, and SVM-based feature ranking.

\noindent\textbf{\ul{\emph{Pathway Performance Integration.}}}
We leverage the KEGG pathway database $P = \{p_1, \dots, p_m\}$ to incorporate biological context. For each pathway $p_i$ containing a gene set $G_i$, we compute a score $S_p(p_i)$ using a classifier trained on genes within that pathway: ${S_p(p_i) = \text{classifier\_performance}(D[G_i])}$, where $D[G_i]$ represents the dataset restricted to genes in pathway $p_i$.

\noindent\textbf{\ul{\emph{Integrative Score Calculation.}}}
The final importance score $\hat{s}_g$ for each gene $g$ combines both statistical and pathway-based evidence. Each method's weight is computed based on its performance:
$
    w_i = \frac{S_p(f_i)}{\sum_{f_j \in F} S_p(f_j)}
$,
where $S_p(f_i)$ is the performance score of method $f_i$ on validation data. The base meta-score for gene $g$ is:
$
    m_g = \sum_i w_i \cdot S_{i,g}
$.
This score is then adjusted using pathway information:
$
    \hat{s}_g = m_g \cdot \left(1 + \beta \cdot \log(1 + \bar{S}_p(g))\right)
$, where $\bar{S}_p(g)$ is the mean performance score of pathways containing gene $g$, and $\beta$ is a scaling factor. Unmapped genes retain their base meta-scores to avoid penalizing potentially important but less studied genes. The final filtered gene set is:
$
    G_{\mathit{pre}} = \{ g \in G : \hat{s}_g > \mu + 2\sigma \}
$
where $\mu$ and $\sigma$ are the mean and standard deviation of the adjusted scores.

\vspace{-0.5cm}
\subsection{Multi-Agent Gene Selection Framework}
\label{stage_2}
\noindent\textbf{Why model genes as collaborative decision-makers?} Building on the pre-filtered gene set $G_{\text{pre}}$, we implement a MARL framework where each gene in $G_{\text{pre}}$ acts as an agent that learns to make selection decisions collaboratively. Traditional feature selection methods treat genes independently, failing to capture the complex interactions within biological pathways. MARL allows us to model genes as collaborative agents to reflect both their individual significance and collective behavior in pathways. Through pathway-aware state representations and a reward mechanism that integrates both statistical and pathway-level performance, agents learn selection strategies that balance predictive power with biological relevance while efficiently handling high-dimensional genomic data.

\noindent\textbf{\ul{\emph{Action Space.}}}
The action space for each agent $i$ (i.e, gene $i$) is binary: $a_i \in \{0, 1\}$, where 0 represents discarding and 1 represents selecting the gene.

\noindent\textbf{\ul{\emph{State Representation.}}}
To effectively capture both gene interactions and pathway relationships, we employ a Graph Neural Network (GNN) for state representation. The state $s_t$ at time step $t$ is constructed as: $s_t = \text{GNN}(X_t, E_t)$, where $X_t$ represents the expression matrix of genes in \( G_{pre} \) augmented with pathway embeddings at time step $t$. The edge connections $E_t$ are constructed through an adaptive mechanism that combines correlation and pathway information:
$
    E_{ij} = \rho \cdot C_{ij} + (1 - \rho) \cdot J_{ij}
$
where $C_{ij}$ is the correlation between genes $i$ and $j$, $J_{ij}$ is the Jaccard similarity of their pathway memberships, and $\rho$ is the correlation weight.
This information is then processed through graph convolutional layers:
\[
    h = \sigma(\tilde{A} \times W^{(0)})
\]
\[
    s_t = \text{pool}(\sigma(\tilde{A} h W^{(1)}))
\]
where $\tilde{A}$ is the normalized adjacency matrix derived from the edge connections, $W^{(0)}$ and $W^{(1)}$ are learnable weight matrices, and $\sigma$ is the ReLU activation function. The pooling operation aggregates node-level features into a global state while preserving both individual gene characteristics and pathway contexts.

\noindent\textbf{\ul{\emph{Multi-Agent Deep $Q$-Network Architecture.}}} Each agent employs a Deep $Q$-Network (DQN) to learn optimal selection strategies. The $Q$-function for agent $i$ (i.e., gene $i$) is approximated by a neural network $Q_i(s, a;\theta _i)$ that maps states to action values, where $\theta _i$ are the network parameters. 
The network learns to estimate the expected cumulative reward for taking action $a$ in state $s$: \[	Q_i (s,a;\theta _i) \approx \mathbb {E}[R_t\mid s_t=s, a_t=a]\]
where $R_t = \sum_{k=0}^\infty \gamma^k r_{t+k}$ is the discounted cumulative reward.
Drastic reduction in feature space while trying to maintain or improve prediction performance does not exactly go hand-in-hand. To stabilize learning and improve efficiency, we implement prioritized experience replay and maintain a target network. Prioritized replay ensures that important transitions are sampled more frequently during training, while the target network, updated periodically, reduces overestimation bias in the $Q$-value updates.
The networks are trained using the Huber loss to provide robustness against outliers, and an epsilon-greedy policy with decaying exploration rate ensures sufficient exploration of the feature space while gradually focusing on exploiting learned strategies.

\noindent\textbf{\ul{\emph{Centralized Critic.}}} 
To provide a global perspective on the collective performance of our multi-agent feature selection system, we implement a centralized critic $V(s)$ that maps the current state $s_t$ to a scalar value estimate $v_t$. The critic architecture combines compression, gating, and dynamic layers to effectively process the high-dimensional state information. It is trained to minimize the mean squared error between its prediction and observed performance improvement: $I_{\text{critic}} = \text{MSE}(v_t, I_t)$, where $I_t$ is global performance improvement at time step $t$.

The critic's value estimates are incorporated into the DQN learning process, influencing the target $Q$-values: $Q_{\text{target}} = \lambda_a (r_t + \gamma \max_a Q(s_{t+1}, a)) + \lambda_b v_t$, where $\lambda_a$ and $\lambda_b$ are weighting factors. This integration helps balance local and global optimization, promoting more stable and effective feature selection by providing a baseline that reduces variance in the policy updates. The global value estimate provided by the critic serves as a form of baseline, helping to reduce variance in the policy gradients implicitly computed by the DQNs through their value function updates.

\noindent\textbf{\ul{\emph{Shared Memory.}}} 
To enhance collaboration between agents and leverage collective knowledge, we introduce a shared memory mechanism consisting of two components: a collaboration success record and a synergy matrix. The collaboration success record $H$ is defined as a mapping from gene sets to their corresponding performance improvements: $H : \mathcal{P}(F) \to \mathsf{R}$ where $F$ is the set of all features, $\mathcal{P}(F)$ power set of $F$, and $\mathsf{R}$ represents performance improvements. For a given feature set $S \in \mathcal{P}(F)$, the collaboration success is updated as $H(S) = \max(H(S), \Delta p)$, where $\Delta p$ is the observed performance improvement when using feature set $S$. The synergy matrix $M \in \mathbb{R}^{d \times d}$, where $d$ is the number of features, captures pairwise feature interactions. For features $i$ and $j$, the synergy value updated as: $M[i,j] \mathrel{+}= \Delta p$, when features $i$ and $j$ are both present in a successful feature combination. Both components follow an exponential decay mechanism to adapt to changing feature relationships over time.

The shared memory influences the action selection process by introducing a synergy bias. The probability of selection action $a$ for feature $i$ is given by: $P(a_i=1) = \sigma \left(Q(s_i,1) + \eta \cdot \sum_{j \in P_i} M[i,j] \right)$, where $Q(s_i,1)$ is the $Q$-value for selecting feature $i$, $P_i$ refers to the top $k$ partners to consider for feature $i$ based on the synergy matrix, \(\sigma\) is the sigmoid function, and \(\eta\) is a hyperparameter controlling the influence of the synergy bias.

\noindent\textbf{\ul{\emph{Reward Mechanism.}}} To address the credit assignment problem in MARL-based feature selection, we implement a sophisticated reward mechanism that combines efficient performance estimation with pathway-level insights to guide agent decisions. The reward calculation consists of three key components:

\noindent\ul{\emph{Personal Performance Impact Estimator.}} First, to estimate individual feature contributions efficiently, we construct a perturbation matrix $D_t$, where each row flips a single element of the selection vector ${a}_t \in \{0,1\}^{d}$, representing an alternative feature state. Evaluating model performance for each state yields labels $y_t$, and $(D_t, y_t)$ is passed as training data to an ensemble meta-learner $f_\text{meta}$ (combining random forest, XGBoost, LightGBM with a meta model and a neural network). Performance changes are estimated as $\Delta {R}_t = f_{\text{meta}}(D_t) - R_t$, where $R_t$ is the current performance. By leveraging matrix-based updates and parallel computation, our approach avoids the inefficiencies of sequential feature evaluations, making it scalable for high-dimensional genomic data.

The reward for each agent incorporates uncertainty-aware adjustments. Given an uncertainty vector $u_t$ estimated by $f_\text{meta}$, we compute the confidence factor ${c}_t = \frac{1}{1 + {u}_t}$ and define the reward as: ${r}_{base} = \Delta {R}_t \odot {{c}_t} - \log(1 + {u}_t) + I_t \mathbf{1}$, where $I_t$ represents global performance improvement. 

To maintain accuracy throughout the selection process, we implement an online learning mechanism. Periodically, at a pre-defined update frequency $F$, we update the meta-learner using a buffer $\mathcal{B} = \{(\mathbf{a}_t, R_t)\}_{t=1}^B$ of recent experiences, where $\mathbf{a}_t$ represents the joint action vector of all agents at time $t$, and $B$ is the buffer size. The update minimizes the loss:
\[L = \sum_{(\mathbf{a}_t, R_t) \in \mathcal{B}} \left(f_\text{meta}(\mathbf{a}_t) - R_t\right)^2 + \lambda \Omega(f_\text{meta})\]
where $\Omega(f_\text{meta})$ is a regularization term with coefficient $\lambda$.

\noindent\ul{\emph{Gene Centrality Measure.}}
We introduce a pathway-based centrality measure that evaluates each gene's importance within and across biological pathways. Unlike traditional centrality metrics that focus solely on network topology, our measure considers both the gene's pathway membership and its connectivity patterns to capture its functional significance.
For a gene $i$, we define a centrality score $\phi_i$ that captures its importance within and across pathways: $\phi_i = n_i \times \sum_{p \in P_i} |G_{i,p}|$, where $n_i$ is the number of pathways containing gene $i$, $P_i$ is the set of pathways containing gene $i$, and $|G_{i,p}|$ represents the number of genes connected to gene $i$ in pathway $p$. 

For a selection $S$, the aggregate centrality function $\Phi(S)$ incorporates both pathway membership and cross-pathway connectivity:

\[
\Phi(S) = \sum_{i \in S} \left(n_i \sum_{p \in P_i} |G_{i,p} \cap S|\right)
\]

For each gene $i$, we calculate a differential centrality score $\Delta\phi_i$ that quantifies how its inclusion or exclusion impacts the overall pathway representation. For unselected genes, we calculate: 
\[
\Delta\phi_i = \Phi(S \cup \{i\}) - \Phi(S)
\]
where $\Phi(S)$ represents the aggregate centrality score for the current selection $S$. For genes already in the selection, we calculate the loss in centrality from their removal:
\[
\Delta\phi_i = \Phi(S) - \Phi(S \setminus \{i\})
\]
This formulation rewards genes that serve as critical connectors between multiple pathways, reflecting their potential regulatory importance in biological processes.

\noindent\ul{\emph{Pathway Coverage Measure.}}
We implement a pathway coverage measure to ensure comprehensive representation of biological processes by evaluating how effectively the selected genes span known biological pathways. For a given pathway \(p\), we define a coverage score \(\psi_p\) that measures the proportion of pathway genes included in the current selection: $\psi_p(S) = \frac{|S \cap G_p|}{|G_p|}$, where \(G_p\) represents the set of genes in pathway \(p\), and \(S\) is the current gene selection. For each gene \(i\), we calculate a differential coverage score \(\Delta\psi_i\) that quantifies its contribution to pathway coverage. For unselected genes, we calculate the potential coverage gain:

\[
\Delta\psi_i = \psi_p(S \cup \{i\}) - \psi_p(S)
\]

For genes already in the selection, we calculate the coverage loss from their removal:

\[
\Delta\psi_i = \psi_p(S) - \psi_p(S \setminus \{i\})
\]
This measure encourages the selection of genes that maximize the representation of diverse biological processes while penalizing redundant selections within already well-covered pathways.



The final reward for each gene $i$ combines the base performance estimate with the pathway-based measures through a weighted sum: $r_i = \omega \cdot r_{base} + \xi \cdot \Delta\phi_i + \zeta \cdot \Delta\psi_i$, where \(\omega, \xi, \zeta\) are weighting coefficients that sum to 1, balancing the importance of statistical performance (\(r_{base}\)), gene centrality impact (\(\Delta\phi_i\)), and pathway coverage contribution (\(\Delta\psi_i\)). 

\noindent\textbf{\ul{\emph{Optimal Set Selection.}}} The final optimal gene set is determined by computing weighted importance scores for each gene based on the $Q$-value differences between its selection and rejection actions over time: $w_i = \sum_{t=1}^T \gamma^{T-t}(Q_i^t(s_t, 1) - Q_i^t(s_t, 0))$, where $\gamma$ is a decay factor giving more weight to recent decisions. Genes are ranked by these scores and the top $k$ genes form the optimal set $G_{opt} = \{g_i \in G_{ranked} : i \leq k\}$ are selected. This captures the cumulative learning of agents while favoring consistently important genes.

\section{Experiment}
\begin{figure*}[!h]
\vspace{-0.2cm}
    \centering  \includegraphics[width=1.0\linewidth,height=8cm,keepaspectratio]{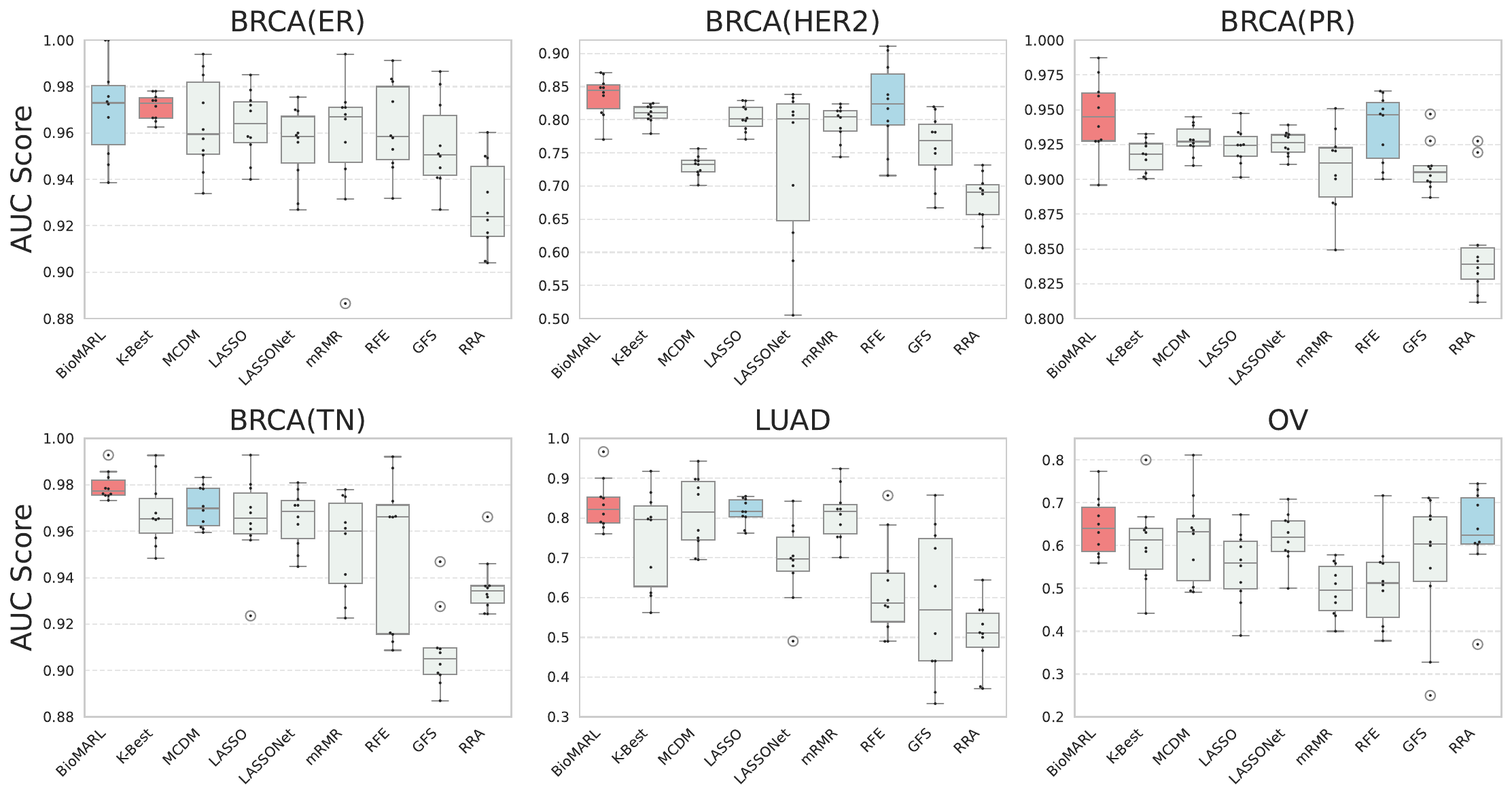}
     \vspace{-0.2cm}
    \captionsetup{justification=centering, font=small}
    \caption{Overall performance comparison of BioMARL with eight state-of-the-art baselines: Best performing baseline highlighted in 
    \highlight{lightcoral}{lightcoral} and second-best in \highlight{lightblue}{lightblue}. 
    }
    \label{fig:performance}
\end{figure*}
\subsection{Experimental Setup.} \label{setup}
\noindent\textbf{Dataset Descriptions.} We performed experiments on the TCGA breast invasive carcinoma (BRCA) \cite{cancer2012comprehensive}, lung adenocarcinoma (LUAD) \cite{cancer2014comprehensive} and ovarian serous cystadenocarcinoma (OV) \cite{cancer2011integrated} datasets. For the BRCA studies, patients were classified based on estrogen receptor (ER+ versus ER-), progesterone receptor (PR+ versus PR-), human epidermal growth factor receptor 2 (HER2+ versus HER2-), and triple negative (TN versus non-TN) status.  For the LUAD and OV studies, we stratified patients into two groups based on survival time: those who survived less than 25 months and those who survived more than 50 months. Each dataset has 20,530 features (i.e., genes). More detailed statistics shown in \textbf{Table \ref{tab:stats_table}}.

\begin{table}[H]
\captionsetup{font=small}
\centering
\small
\vspace{-0.2cm}
\setlength{\tabcolsep}{3pt} 
\caption{Sample distribution across datasets.  
Class 1: ER+, HER2+, PR+, TN, survival time $<$25 months.  
Class 2: ER-, HER2-, PR-, non-TN, survival time $>$50 months.}
\vspace{-0.3cm}
\begin{tabular}{@{}c|cc@{}}
\toprule
Dataset      & Class 1 & Class 2 \\ \midrule
BRCA (ER)   & 332     & 80      \\
BRCA (HER2) & 334     & 78      \\
BRCA (PR)   & 285     & 127     \\
BRCA (TN)   & 65     & 347      \\
LUAD        & 109     & 21      \\
OV          & 63      & 49      \\ \bottomrule
\end{tabular}
\vspace{-0.2cm}
\label{tab:stats_table}
\end{table}

\noindent\textbf{Baseline Algorithms.}
We compared BioMARL with eight widely used feature selection methods: 1. \textbf{K-Best}~\cite{yang1997comparative} selects K features with the highest feature score; 2. \textbf{mRMR} ~\cite{peng2005feature} selects a feature subset with the highest relevance to the target and least redundancy among themselves;
3. \textbf{LASSO} ~\cite{tibshirani1996regression} uses regularization to shrink coefficients of less useful features to zero,  effectively performing feature selection during model fitting; \
4. \textbf{RFE} ~\cite{guyon2002gene} recursively removes the weakest features until a specified number of features is reached;
5. \textbf{LASSONet} ~\cite{lemhadri2021lassonet} is a neural network with sparsity to encourage the network to use only a subset of input features;
6. \textbf{GFS} ~\cite{babatunde2014genetic} selects features using genetic algorithms, which recursively generates a population based on a possible feature subset, then uses a predictive model to evaluate it; 
7. \textbf{RRA} ~\cite{seijo2017ensemble} collects distinct selected feature subsets, and then integrates them based on statistical sorting distribution;
8. \textbf{MCDM} ~\cite{hashemi2022ensemble} approaches feature selection as a Multi-Criteria Decision-Making problem and uses the VIKOR sort algorithm to rank features based on the judgment of multiple selection methods. \ul{All experiments use AUC score as the metric to evaluate the classification performance.} The baselines along with our method, were executed 10 times on each dataset for every experiment, and the average AUC score across these runs was reported for each method. We adopted Random Forest as the downstream classification model and during each run, a 70-30 training-testing split was employed, with genomic feature selection performed using 5-fold cross-validation on the training set, followed by evaluation of the selected features on the hold-out set.  All reported performance scores correspond to $k=100$ selected features.

\vspace{-0.2cm}
\subsection{Experimental Results}
\noindent\textbf{Overall Comparison.} 
In this section, we assess the performance of BioMARL and baseline algorithms for feature selection on the aforementioned datasets. Figure \ref{fig:performance} shows the comparison result. BioMARL outperforms other selection algorithms in 5 out of the 6 datasets, demonstrating consistently higher AUC scores. Additionally, it exhibits relatively low variance in most cases, indicating stable and reliable performance. The underlying driver behind this performance is BioMARL's dual-layer optimization approach. The pathway-guided pre-filtering first creates a high-quality candidate pool by eliminating statistically insignificant genes while preserving meaningful pathway relationships. This provides a strong foundation for feature selection which the subsequent MARL framework then leverages through sophisticated reward estimation that considers both individual gene contributions and their synergistic effects, allowing agents to identify feature combinations that maximize classification performance.

\begin{table*}[]
\centering
\captionsetup{font=small}
\caption{Ablation study of BioMARL. Best performance highlighted in \textbf{bold}}
\vspace{-0.1cm}
\label{tab:ablation}
\resizebox{\textwidth}{!}{
\begin{tabular}{c|ccc|cccc}
\hline
\multirow{2}{*}{Variant} & \multicolumn{3}{c|}{Technical Component} & \multicolumn{4}{c}{Performance} \\ \cline{2-8} 
 & \multicolumn{1}{c|}{Personalized Reward} & \multicolumn{1}{c|}{Centralized Critic} & Shared Memory & \multicolumn{1}{c|}{BRCA(ER)} & \multicolumn{1}{c|}{BRCA(HER2)} & \multicolumn{1}{c|}{LUAD} & OV \\ \hline
$\text{BioMARL}^{-\text{Rwd}}$ & \xmark & \cmark & \cmark & 0.9598 $\pm$ 0.023 & 0.8184 $\pm$ 0.061 & 0.7064 $\pm$ 0.107 & 0.6276 $\pm$ 0.106 \\
$\text{BioMARL}^{-\text{Crt}}$ & \cmark & \xmark & \cmark & 0.9605 $\pm$ 0.029 & 0.8073 $\pm$ 0.030 & 0.7623 $\pm$ 0.058 & 0.6298 $\pm$ 0.095 \\
$\text{BioMARL}^{-\text{Mem}}$ & \cmark & \cmark & \xmark & 0.9618 $\pm$ 0.0166 & 0.7914 $\pm$ 0.093 & 0.7712 $\pm$ 0.089 & \textbf{0.6494} $\pm$ \textbf{0.100} \\
BioMARL & \cmark & \cmark & \cmark & \textbf{0.9706} $\pm$ \textbf{0.0198} & \textbf{0.8357} $\pm$ \textbf{0.0296} & \textbf{0.8327} $\pm$ \textbf{0.0601} & 0.6439 $\pm$ 0.065 \\ \hline
\end{tabular}
}
\end{table*}


\noindent\textbf{Ablation Study.} To validate the impact of each technical component, we developed three variants of BioMARL: (i) \textbf{BioMARL$^{-Rwd}$} uses performance improvement directly as the reward for RL agents, (ii) \textbf{BioMARL$^{-Crt}$} operates without a centralized critic, and (iii) \textbf{BioMARL$^{-Mem}$} excludes the shared memory mechanism. Table \ref{tab:ablation} presents the comparison results for two breast cancer studies, as well as the LUAD and OV datasets. BioMARL outperforms \textbf{BioMARL$^{-Rwd}$}, demonstrating that the pathway-aware reward mechanism effectively balances statistical significance with biological relevance, resulting in more robust and biologically meaningful feature selection. The superior performance over \textbf{BioMARL$^{-Crt}$} highlights the importance of centralized critic in capturing global state information, resulting in better-coordinated agent behaviors and avoiding conflicting local optimizations. Finally, advantage over \textbf{BioMARL$^{-Mem}$} demonstrates the value of shared memory mechanism, which enables collaborative learning and accelerates the discovery of synergistic feature sets by avoiding redundant exploration of suboptimal combinations. 

\begin{figure}[]
\vspace{-0.2cm}
    \centering
    \includegraphics[width=1.0\linewidth]{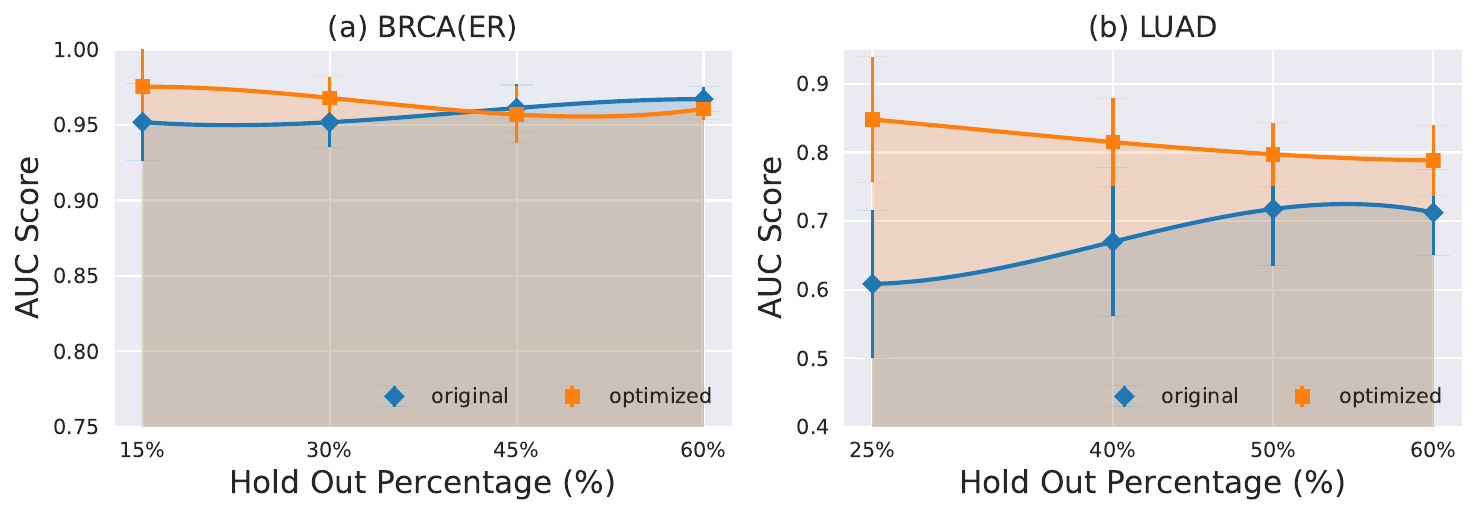}
    \vspace{-0.2cm}
    \captionsetup{justification=centering, font=small}
    \vspace{-0.33cm}
    \caption{The influence of different hold-out percentages in terms of AUC score. `optimized' refers to the subset of features selected by BioMARL.}
    \vspace{-0.5cm}
    \label{fig:hold_out}
\end{figure}

\noindent\textbf{Study of the Impact of Hold-out Percentage.}  \label{hold-out}
In gene selection problems with imbalanced datasets and limited samples, evaluating feature selection algorithms across various train-test splits, including unconventional ratios, might be beneficial. This highlights the model's robustness in scenarios with smaller sample sizes, where traditional splits (80-20 or 70-30) may not suffice. Additionally, it offers practical insights into model performance when data availability is limited, which is common in expensive and data-scarce applications.  In the experiment setup section, we used a 70-30 train-test split. We further explored the impact of the holdout setting through experiments on BRCA(ER) and LUAD, as shown in Figure \ref{fig:hold_out}. While they exhibit varying patterns due to factors like randomness of dataset partitioning, training data volume, and task complexity, BioMARL consistently demonstrates performance enhancements. Despite the lack of uniform trend across them, the results generally indicate that BioMARL reliably improves performance across most scenarios. This consistency affirms the robustness of our approach regardless of the holdout percentage used.

\noindent\textbf{Clear Separation of Breast Cancer Patients by BioMARL-Selected Genes.}  \label{heatmap}
The heatmap of gene expression data for top 100 genes selected by BioMARL exhibits a well-defined pattern, achieving perfect separation between PR+ and PR- breast cancer patients. Most of the selected genes show elevated expression in PR- patients, suggesting their involvement in pathways related to PR signaling and breast cancer progression. This clear distinction underscores BioMARL's effectiveness in identifying biologically meaningful and clinically relevant features, reinforcing its potential as a powerful tool for biomarker discovery. Identifying genes that distinguish PR subtypes is critical for understanding the molecular mechanisms driving breast cancer heterogeneity and could provide valuable insights for prognosis and targeted therapy development. Notably, similar clear expression patterns were also observed in other breast cancer subtypes which further validate BioMARL’s robustness in capturing key regulatory signatures, highlighting its significance in advancing precision oncology and personalized treatment. A potential reason behind this is its collaborative multi-agent architecture that enables agents to identify complementary genes that collectively maximize class separation, rather than relying on individual significance, forming robust discriminative patterns.

\begin{figure}[!h]
\vspace{-0.2cm}
    \centering
    \includegraphics[width=1.0\linewidth]{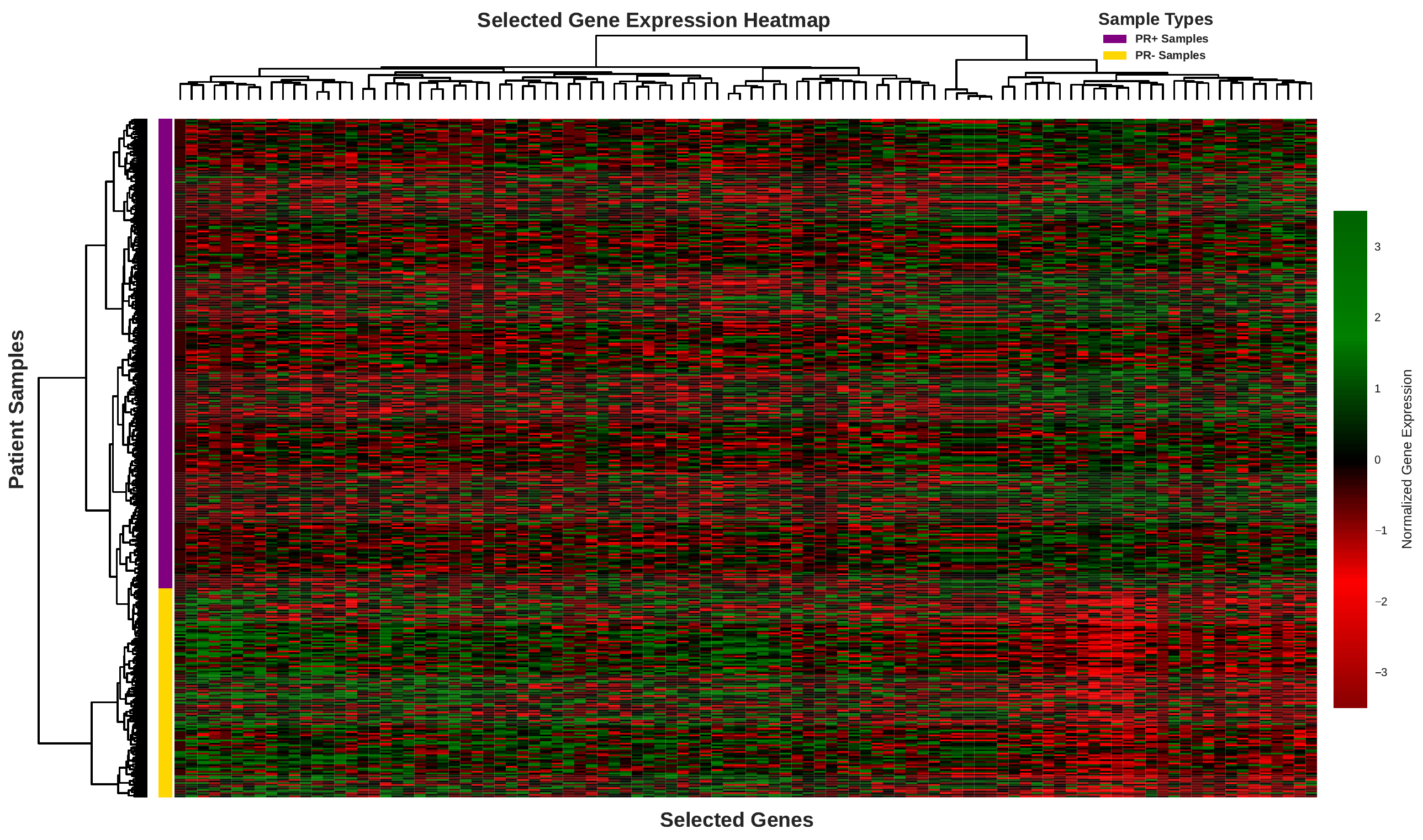}
    \vspace{-0.2cm}
    \captionsetup{justification=centering, font=small}
    \vspace{-0.33cm}
    \caption{Heatmap of the expression profiles of 100 marker genes selected by BioMARL on the BRCA(PR) dataset.}
    \vspace{-0.2cm}
    \label{fig:heatmap}
\end{figure}

\begin{figure}[!h]
\vspace{-0.2cm}
    \centering
    \includegraphics[width=1.0\linewidth]{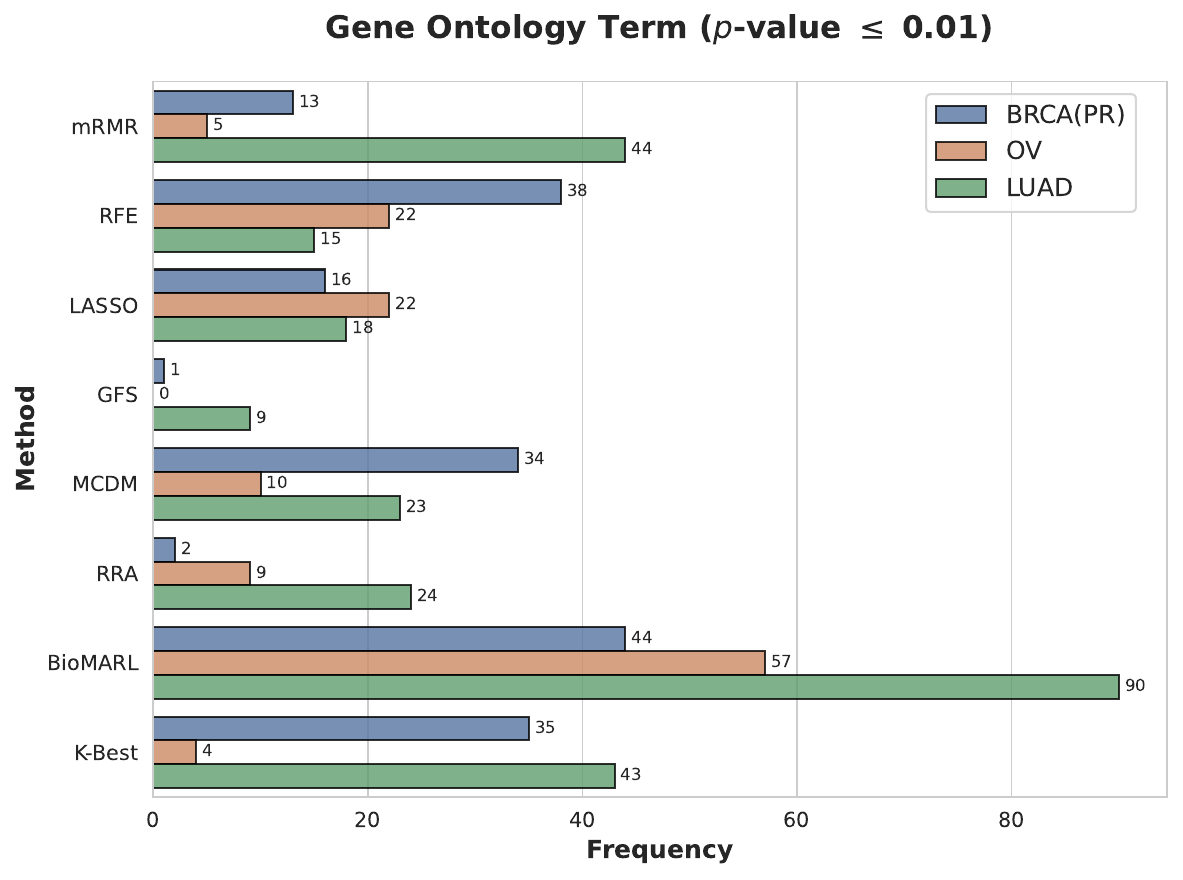}
    \vspace{-0.2cm}
    \captionsetup{justification=centering, font=small}
    \vspace{-0.33cm}
    \caption{Enrichment analysis of selected genes for breast, ovarian and lung cancer datasets. Bar plots show the frequency of significantly enriched Gene Ontology terms ($p$-value $\leq$ 0.01) for different selection methods.}
    \vspace{-0.3cm}
\label{fig: gene_enrichment}
\end{figure}

\noindent\textbf{Gene Enrichment Analysis.}To evaluate the biological relevance of genes selected by BioMARL and baseline methods, we performed gene set enrichment analysis. This analysis aims to identify significant overlaps between selected genes and known functional gene sets, which may provide insights into the biological processes and pathways potentially involved in these cancers. We examined the enrichment of Gene Ontology (GO) terms \cite{ashburner2000gene} among the top-ranked 100 genes identified by each method. Figure \ref{fig: gene_enrichment}  illustrates the frequency of enriched GO terms $p$-value $\leq$ 0.01 for all different methods across three datasets. BioMARL demonstrated superior performance on both datasets which suggests that it effectively captures genes involved in biologically relevant processes specific to different cancer types. The underlying impetus for this is BioMARL's pathway-guided pre-filtering that establishes a biologically relevant initial feature space, and its ability to combine biological domain knowledge with collaborative learning, where intelligent agents make selection decisions guided by both pathway information and collective performance optimization.

\begin{figure}[]
\vspace{-0.2cm}
    \centering
    \includegraphics[width=0.7\linewidth]{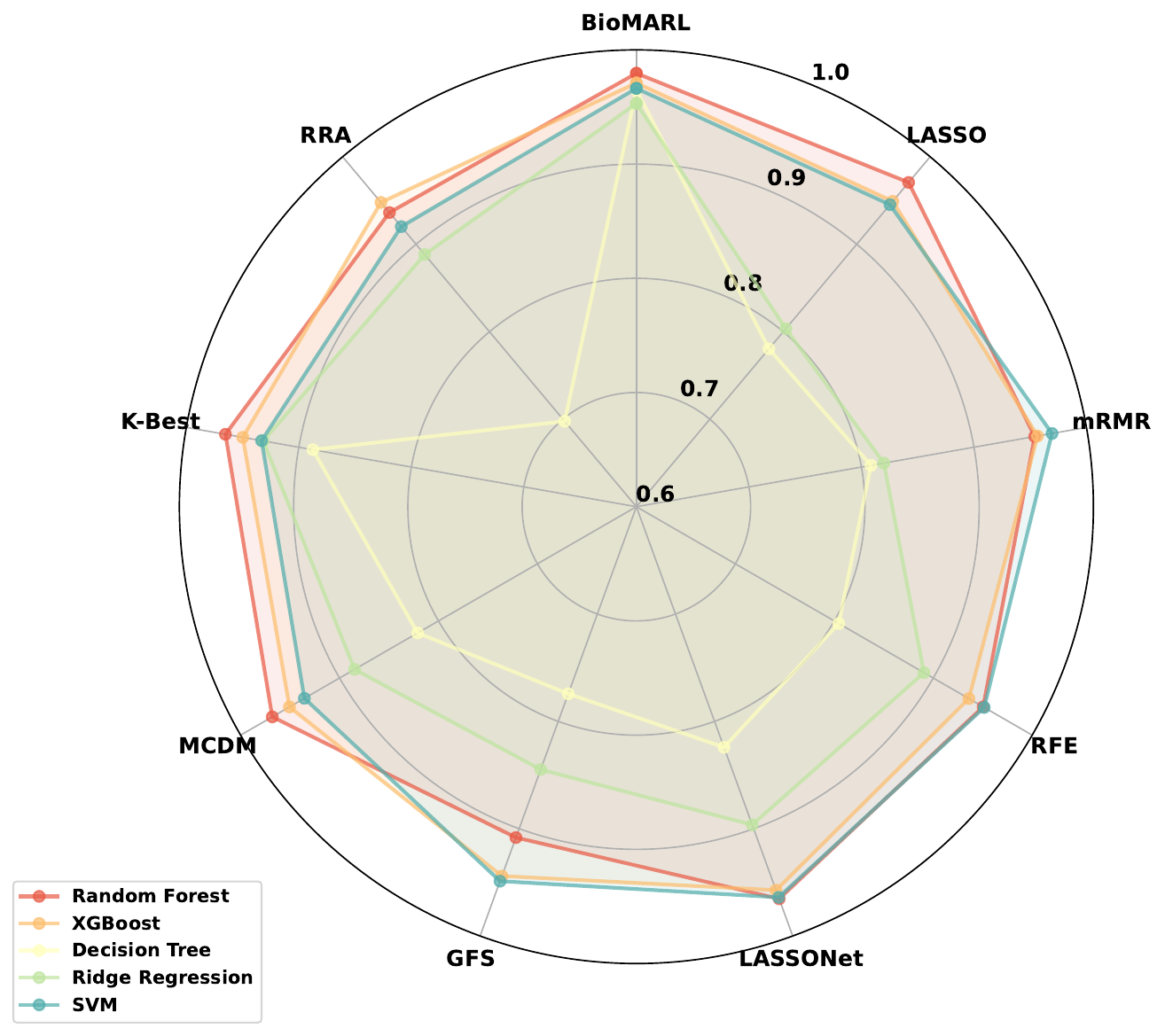}
    \vspace{-0.2cm}
    \captionsetup{font= small, justification=centering, font = small}
    \vspace{-0.1cm}
    \caption{Robustness evaluation of BioMARL across classification models, measured by AUC score on BRCA(TN). Larger area coverage indicates better overall performance across feature selection methods.}
    \vspace{-0.5cm}
\label{fig: robustness}
\end{figure}

\noindent\textbf{Robustness Check of BioMARL Under Different Classification Models.} Feature selection methods depend on evaluating the currently selected feature subset at each step with a downstream predictive model to identify the optimal feature subset. In our design, we employed the Random Forest for evaluation due to its stability and robustness. However, more rigorously, a feature selection method should be able to generalize to diverse downstream models. We examine the robustness of BioMARL on BRCA(TN) by changing the classifier to XGBoost, SVM, Decision Tree, LASSO and Ridge Regression. The comparison results, depicted in Figure \ref{fig: robustness} shows  that BioMARL exhibits downstream model-agnostic robustness and reliability. A potential reason behind this robustness is our multi-agent framework's ability to learn feature importance through collaborative decision-making and shared memory, rather than being tied to any specific model's optimization criteria. The agents learn selection strategies that identify fundamentally predictive features through their collective experience, leading to stable performance regardless of the downstream classifier used.

\begin{table*}[htbp]
\centering
\caption{Literature review of the candidate cancer genes. This table presents citations highlighting the relevance of the genes selected exclusively by BioMARL.}
\vspace{-0.2cm}
\label{tab:lit_review}
\resizebox{\textwidth}{!}{
\begin{tabular}{cccc}
\hline
\textbf{Gene Name} & \textbf{Pathway} & \textbf{Desccription} & \textbf{Reference} \\ \hline
\rowcolor{gray!30}
\textit{CYP3A7} & hsa00140  & \begin{tabular}[c]{@{}c@{}}When expressed in adults, driven by the \textit{CYP3A7*1C} allele, \\ is associated with altered hormone metabolism and impacts breast cancer outcomes.\\\end{tabular} & ~\cite{johnson2016cytochrome} \\
\textit{NME1} & hsa00230 & \begin{tabular}[c]{@{}c@{}} Upregulated in DCIS but downregulated in invasive breast cancer;\\ loss of \textit{NME1} promotes tumor invasion by increasing \textit{MT1-MMP} surface levels.\end{tabular} & ~\cite{lodillinsky2021metastasis} \\
\rowcolor{gray!30}
\textit{FGA} & hsa04610 & \begin{tabular}[c]{@{}c@{}}Identified as a key biomarker in HER2-positive breast cancer plasma samples, \\ with lower levels in cancer patients compared to controls, reverting to control levels post-surgery. \end{tabular} & ~\cite{wang2020fibrinogen} \\
\textit{GRB7} & hsa04140  & \begin{tabular}[c]{@{}c@{}}Co-amplified with HER2 gene in breast cancer, GRB7 facilitates HER2-mediated signaling\\ and tumor formation, with its protein overexpression concurrent with HER2 amplification.\end{tabular} & ~\cite{bivin2017grb7} \\
\rowcolor{gray!30}
\textit{ERBB2} & hsa04520 & \begin{tabular}[c]{@{}c@{}}Major oncogene amplified in 20-30\% of breast cancers, \\ its overexpression correlates with tumor chemoresistance and poor patient prognosis,\\ making it a crucial therapeutic target. \end{tabular} & ~\cite{fernandez2022examination} \\
\textit{AGTR1} & hsa04614 & \begin{tabular}[c]{@{}c@{}} Shows profound overexpression in a subset of breast tumors across independent cohorts, \\ suggesting a potential role in tumorigenesis and as a therapeutic target.\end{tabular}  & ~\cite{rhodes2009agtr1} \\
\rowcolor{gray!30}
\textit{PSMC1} & hsa03050 & \begin{tabular}[c]{@{}c@{}}Significantly upregulated in breast cancer tissues, with high expression correlating with\\ poor survival outcomes and involvement in critical cancer-related cellular processes.\end{tabular}  &  ~\cite{kao2021prognoses} \\
\textit{TUBB3} & hsa04540 & \begin{tabular}[c]{@{}c@{}}High TUBB3 expression in breast cancer correlates with decreased sensitivity\\ to taxane-based chemotherapy and is associated with high tumor grade and advanced tumor stage. \end{tabular}  & ~\cite{lebok2016high} \\\hline
\end{tabular}
}
\end{table*}
\noindent\textbf{Case Study \uppercase\expandafter{\romannumeral 1}: BioMARL Selected Genes Have Proven Relevance in Cancer.} To further validate the biological relevance of genes uniquely selected by BioMARL, we conducted a literature review focusing on these genes in the context of breast cancer. Table \ref{tab:lit_review} highlights a subset of the top-ranked genes and their known associations with breast cancer pathogenesis. Our analysis revealed that several of these genes have established roles in breast cancer progression, metastasis, or treatment response. For instance, the adult expression of \textit{CYP3A7} is linked to altered estrogen metabolism, potentially influencing breast cancer outcomes and response to specific chemotherapeutic agents \cite{johnson2016cytochrome}. \textit{ERBB2} is amplified in 20–30\% of breast cancers, driving chemoresistance and poor prognosis, making its overexpression a key therapeutic target \cite{fernandez2022examination}. \textit{GRB7} not only facilitates HER2-mediated signaling and tumor formation in breast cancer cells but can also be selectively retained and overexpressed in some solid tumors independent of HER2 amplification \cite{bivin2017grb7}. Additionally, \textit{PSMC1} exhibits significantly higher expression in breast cancer tissues compared to normal tissues, with high transcript levels correlating with poor survival outcomes, suggesting its potential as a prognostic biomarker for aggressive disease \cite{kao2021prognoses}.

\begin{figure}[!h]
    \vspace{-0.4cm}
    \centering
    \subfigure[BRCA(ER)]{\label{exp:graphic-memory}\includegraphics[width=0.235\textwidth]{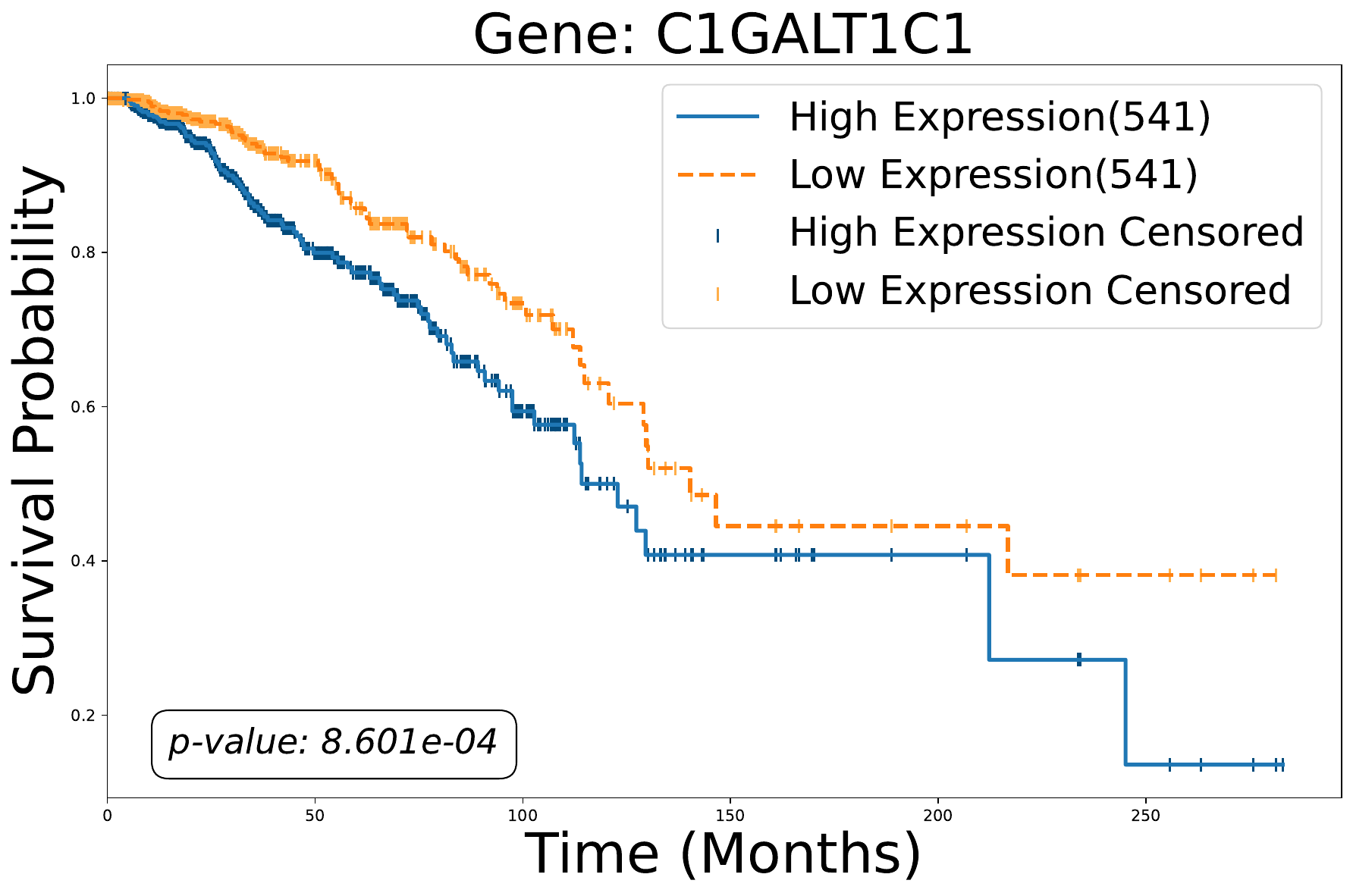}}
    \subfigure[LUAD]{\label{exp:converge-speed}\includegraphics[width=0.235\textwidth]{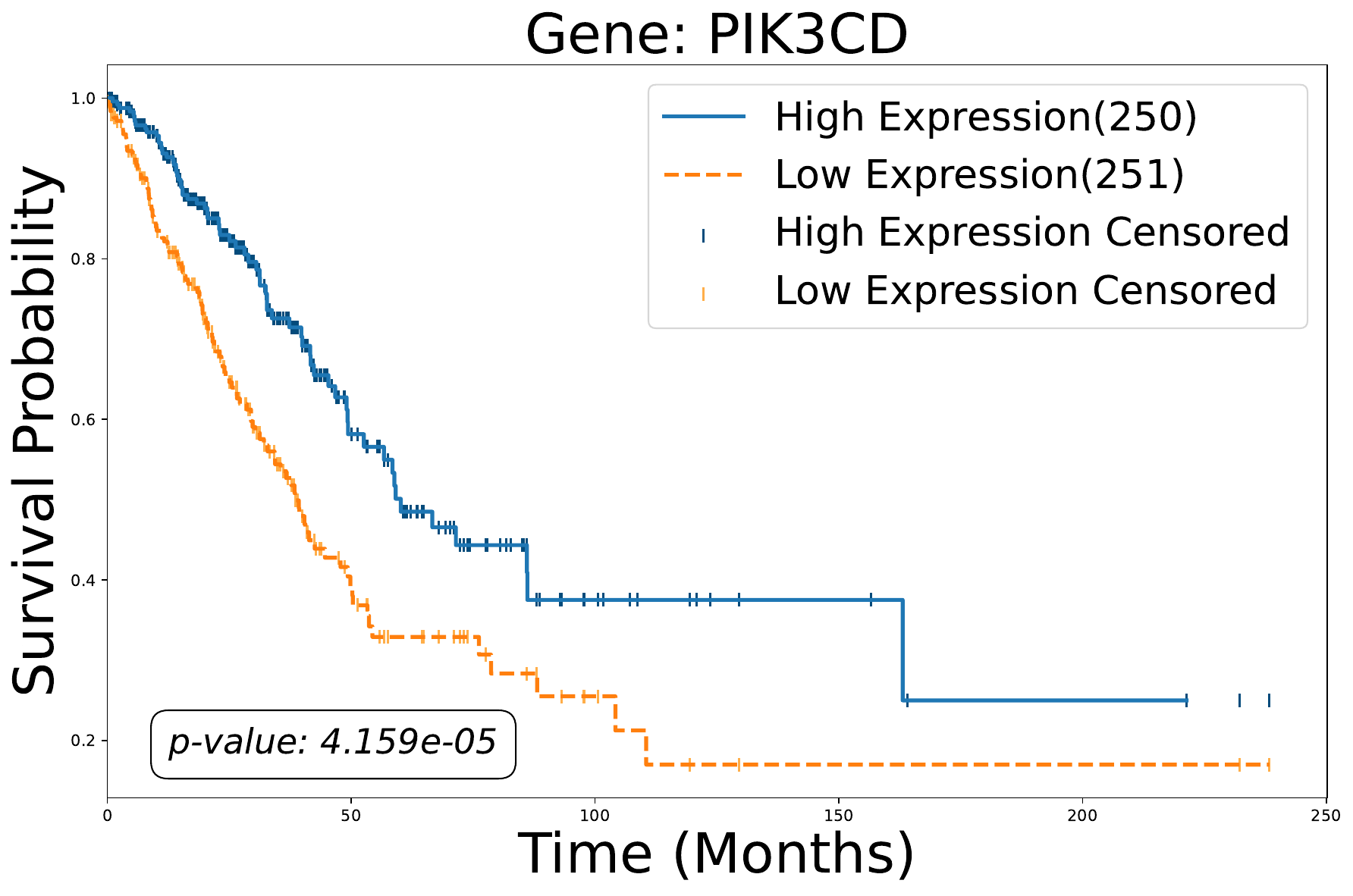}}
    \vspace{-0.4cm}
    \caption{Survival analysis on (a) breast cancer and (b) lung cancer patients with gene selected by BioMARL}
    \vspace{-0.2cm}
    \label{km_curves}
\end{figure}

\noindent\textbf{Case Study \uppercase\expandafter{\romannumeral 2}: BioMARL Improved Survival Prediction.} To evaluate the prognostic significance of the top genes selected by BioMARL, we conducted a survival analysis on cancer datasets. Patients' overall survival was assessed using Kaplan-Meier (KM) plots, stratified by high and low expression levels of the top-ranked genes identified by BioMARL. As shown in Figure \ref{km_curves}, KM survival curves were generated for both groups, and the log-rank test was performed to compare their survival distributions. The KM curves revealed a clear separation between the two groups for the selected genes, indicating potential prognostic value. Additionally, the log-rank test produced significant $p$-values for these genes, reinforcing their association with overall survival in different cancer patients. This analysis demonstrates that genes selected by BioMARL are not only relevant to the disease but also have significant implications for patient outcomes. 

\noindent\textbf{Real-World Deployment and Validation of BioMARL in Gastric Cancer Patient Data.}
Validating a single gene as a cancer biomarker requires substantial financial investment, as biologists and biomedical scientists must conduct gene knockout or knockdown experiments to assess its effects on cell viability, proliferation, and apoptosis \textit{in vitro}. These studies are typically performed in mouse models or cancer cell lines, rather than in actual cancer patients. Identifying a viable drug target and developing a therapeutic for real patients requires clinical trials, which are essential for evaluating safety and efficacy. Over the past two decades, the average cost of drug development has ranged from \$539 million to \$2.8 billion \cite{mulcahy2025use}. The proposed BioMARL can help prioritize target genes, significantly reducing wet-lab experiment costs and lowering research and development (R\&D) expenses in drug discovery.

To evaluate the real-world applicability of BioMARL, we analyzed cancer patient data from Vanderbilt University Medical Center. Compared to the widely used $t$-test, a fundamental statistical method for biomarker identification in cancer studies \cite{van2002gene}, BioMARL-selected genes demonstrated stronger biological relevance, with significant associations with key cancer pathways and greater therapeutic potential, as evidenced by their prevalence in high-impact cancer journals. These results demonstrate the robustness of BioMARL and highlight its potential to advance biomarker-driven precision oncology by improving drug development and clinical trial design.

\vspace{-0.2cm}
\section{Related Work} 

\noindent\textbf{Gene Selection in Genomic Studies.} Feature engineering algorithms for gene expression data fall into three categories: filter, wrapper, and embedded methods. Filter methods~\cite{park2019wx, garcia2020unsupervised, gakii2022graph} evaluate and filter features based on their potential contribution to model performance, typically using relevance scores and thresholds for gene selection. For example, the Grouping Genetic Algorithm~\cite{garcia2020unsupervised} addresses feature grouping in RNA-Seq data. Wrapper methods~\cite{li2017comprehensive,al2022biomarker} assess feature significance using classification algorithms like k-Nearest Neighbors, Random Forests, and Support Vector Machines. Embedded methods~\cite{zhang2021robust, kong2018graph} combine classifier configuration with feature subset exploration. However, these traditional approaches often neglect crucial gene interactions and biological context, while struggling to handle the dynamic nature of gene expression data, potentially missing key biomarkers.

\noindent\textbf{Multi-agent Reinforcement Learning (MARL).} MARL addresses multi-agent environments~\cite{tampuu2017multiagent} where agents interact to achieve shared or individual goals. The combinatorial nature and environmental complexity~\cite{liu2021multi} make training challenging, with many MARL applications addressing NP-Hard problems like manufacturing scheduling~\cite{dittrich2020cooperative, gabel2007successful}, vehicle routing~\cite{silva2019reinforcement, zhang2020multi}, and multi-agent games~\cite{bard2020hanabi, peng2017multiagent}. While deep RL has advanced to handle large-scale systems~\cite{yang2018mean}, and both multi-agent~\cite{ying2024feature, liu2019automating, azim2024feature, xiao2023beyond, gong2025neuro, ying2024revolutionizing} and single-agent~\cite{wang2024knockoff, zhao2020simplifying} approaches have been developed for feature selection, these methods lack comprehensive domain knowledge integration and are not optimized for HDLSS data scenarios, where direct application without proper dimensionality reduction can lead to computational inefficiency and poor convergence.

\vspace{-0.2cm}
\section{Concluding Remarks}
We introduce \textit{BioMARL}, a novel framework that transforms gene selection into a pathway-aware collaborative decision process for genomic analysis. By unifying statistical feature selection with biological pathway knowledge through MARL, our approach bridges the divide between purely statistical methods and biological context in gene selection. The framework's two-stage architecture, combining pathway-guided pre-filtering with collaborative agent modeling, ensures both statistical rigor and biological relevance in selected gene signatures. Through extensive evaluation on multiple TCGA datasets, BioMARL demonstrates superior predictive performance while maintaining interpretability through pathway-level insights. Gene Ontology enrichment analysis confirms that selected genes form biologically coherent signatures, validating our framework's ability to capture meaningful pathway interactions. These genes exhibit distinct expression patterns that differentiate breast cancer patients, highlighting the framework’s potential for biological discovery and its applicability in precision oncology.

\section{Acknowledgement}
{This research was partially supported by the National Science Foundation (NSF) via the grant numbers:  2152030, 2246796, 2426340, 2416727, 2421864, 2421865, 2421803, and National Academy of Engineering Grainger Foundation Frontiers of Engineering Grants.
}

\bibliographystyle{ACM-Reference-Format}
\balance
\bibliography{refs}

\section{Appendix}

\subsection{Implementation Details}
\noindent\textbf{Hyperparameters and Reproducibility} 

\noindent\ul{\textit{Pathway-Guided Gene Pre-filtering:}}
For each pathway, we evaluate performance using a Random Forest classifier 
with 100 trees and default scikit-learn parameters. The pathway performance 
bonus scaling factor $\beta$ is set to 0.2.

\noindent\ul{\textit{Multi-Agent Gene Selection Framework:}}
Each agent's DQN consists of a 4-layer network (256-128-64-2 neurons) with ReLU 
activation and layer normalization. The learning rate is 0.0003, the discount factor 
$\gamma$ is 0.85, the initial $\epsilon$ is 0.95 with a decay rate of 0.99, and the minimum $\epsilon$ is 0.1. 
The experience replay buffer capacity is 1700 with 3000 exploration steps for each run, with a batch size of 64. 
Target network updates occur every 50 steps. The Graph Neural Network uses two graph convolutional layers with a hidden dimension of 64, 
and pathway embeddings of dimension 64. Edge creation uses a correlation weight $\alpha$ of 0.7. $\lambda_a$ and $\lambda_b$ of the critic was set to 0.7 and 0.3 respectively. The meta-learner combines Random Forest, XGBoost, and LightGBM regressors (each with 100 estimators), 
with a linear meta-model and a neural network (256-128-64-32-1 neurons)..
The final reward weighting coefficients $(\omega, \xi, \zeta)$ were set to 0.5, 0.25, and 0.25 respectively 
Online updates of the meta leaner occur every 50 steps.
The shared memory mechanism uses a decay factor of 0.99 and an exploration bonus of 0.005. 
The synergy bias weight $\beta$ is initialized at 0.08 and increases linearly up to 0.3. All experiments were repeated 10 times with different random seeds.

\noindent\textbf{Environmental Settings.} All experiments are conducted on the Ubuntu 22.04.5 LTS operating system, AMD Ryzen Threadripper 2950X 16-Core Processor, and 3 NVIDIA RTX A4500 GPUs (20GB VRAM each) with 128GB of system RAM. The framework uses Python 3.10.12 and PyTorch 2.2.1 with CUDA 11.8 support.

\begin{algorithm}[]
\small
\caption{Pathway-Aware Gene Selection Framework}
\label{pseudo-code}
\begin{algorithmic}[1]
\Require Dataset $D$ with gene set $G$, KEGG pathway database $P$, Maximum features $k$
\Ensure Selected gene set $G_{opt}$ with $|G_{opt}| = k$

\State \textbf{Phase 1: Pathway-Guided Pre-filtering (Section \ref{stage_1})}
\ForAll{method $f_i \in \{$chi-squared, random forest, SVM$\}$}
    \State Compute scores $S_i \gets f_i(D)$ for all genes
    \State $w_i \gets \text{performance}_i/\sum_j \text{performance}_j$ 
\EndFor
\ForAll{pathway $p_i \in P$}
    \State $S_p(p_i) \gets \text{classifier\_performance}(D[G_i])$
\EndFor
\ForAll{gene $g \in G$}
    \State $m_g \gets \sum_i w_i \cdot S_{i,g}$ 
    \State $\hat{s}_g \gets m_g \cdot (1 + \beta \cdot \log(1 + \bar{S}_p(g)))$
\EndFor
\State $G_{pre} \gets \{g \in G : \hat{s}_g > \mu + 2\sigma\}$

\State \textbf{Phase 2: Multi-Agent Gene Selection (Section \ref{stage_2})}
\State Initialize DQN agents for each gene in $G_{pre}$
\State Initialize shared memory $M$ with synergy matrix $M_{ij}$ and success record $C$
\State Initialize centralized critic $V$ with gating mechanism
\State Initialize meta-learner ensemble $f_{meta}$
\For{episode $= 1$ to $E$}
    \State $s_t \gets \text{GNN}(X_t, E_t)$ with pathway-augmented edges
    \For{$t = 1$ to $T$}
        \ForAll{agent $i$}
            \State Get top partners $P_i$ from $M_{ij}$
            \State $\text{bias}_i \gets \beta \sum_{j \in P_i} M_{ij}$
            \State $P(a_i=1) \gets \sigma(Q_i(s_t,1) + \text{bias}_i)$
            \State Select $a_i$ using $\epsilon$-greedy policy
        \EndFor
        
        \State // Performance estimation
        \State $D_t \gets$ construct perturbation matrix
        \State $\Delta P_t, u_t \gets f_{meta}(D_t)$ \Comment{Changes and uncertainties}
        \State $c_t \gets 1/(1 + u_t)$ \Comment{Confidence factors}
        \State $r_{base} \gets \Delta P_t \odot c_t - \log(1 + u_t)$
        
        \State // Pathway metrics
        \ForAll{gene $i$}
            \If{$a_i = 0$}
                \State $\Delta\phi_i \gets \Phi(S \cup \{i\}) - \Phi(S)$
                \State $\Delta\psi_i \gets \psi_p(S \cup \{i\}) - \psi_p(S)$
            \Else
                \State $\Delta\phi_i \gets \Phi(S) - \Phi(S \setminus \{i\})$
                \State $\Delta\psi_i \gets \psi_p(S) - \psi_p(S \setminus \{i\})$
            \EndIf
            \State $r_i \gets \alpha r_{base} + \beta \Delta\phi_i + \gamma \Delta\psi_i$
        \EndFor
        
        \State // Critic and memory updates
        \State $v_t \gets V(s_t)$ \Comment{Global value estimate}
        \State Update $M_{ij}$ for selected gene pairs
        \State Decay memory weights: $M_{ij} \gets \lambda M_{ij}$
        
        \ForAll{agent $i$}
            \State $Q_{target} \gets \lambda_a(r_i + \gamma \max_{a'} Q_i(s_{t+1},a')) + \lambda_b v_t$
            \State Update DQN using $(s_t, a_i, r_i, s_{t+1}, Q_{target})$
        \EndFor
        \State Update critic by minimizing $\|v_t - I_t\|^2$
        \State $s_t \gets s_{t+1}$
    \EndFor
\EndFor

\State Compute final weighted scores using $Q$-value differences
\State Rank genes based on weighted scores to get $G_{ranked}$
\State $G_{opt} \gets$ top $k$ genes from $G_{ranked}$
\State \Return $G_{opt}$
\end{algorithmic}
\end{algorithm}

\subsection{Pseudocode of the algorithm}

The entire procedure algorithm is described as Algorithm~\ref{pseudo-code}.

\end{document}